\documentclass[10pt]{article} % For LaTeX2e
\usepackage[preprint]{tmlr}

\makeatletter
\renewcommand{\headrulewidth}{0pt}
\makeatother

\IfFileExists{math_commands.tex}{%%%%% NEW MATH DEFINITIONS %%%%%

\usepackage{amsmath,amsfonts,bm}

\def\eqref#1{equation~\ref{#1}}
\def\1{\bm{1}}

\DeclareMathAlphabet{\mathsfit}{\encodingdefault}{\sfdefault}{m}{sl}
\SetMathAlphabet{\mathsfit}{bold}{\encodingdefault}{\sfdefault}{bx}{n}

}{}

\usepackage{hyperref}
\hypersetup{hidelinks}
\usepackage{url}
\usepackage{graphicx}
\usepackage{amsmath, amssymb}
\usepackage{amsthm}
\usepackage{booktabs}
\usepackage{xspace}
\usepackage{array}
\usepackage{longtable}
\usepackage{tabularx}
\usepackage{tikz}
\usetikzlibrary{arrows.meta,positioning}
\usepackage{enumitem}
\usepackage{fontawesome5}

\renewcommand{\arraystretch}{1.08}

\usepackage[most]{tcolorbox}

\definecolor{wfblue}{HTML}{2F6F8F}
\definecolor{acgindigo}{HTML}{4E5D94}
\definecolor{tripurple}{HTML}{7A4E8A}
\definecolor{gdtorange}{HTML}{B06A1F}
\definecolor{scopegreen}{HTML}{4C7A48}
\definecolor{glossbg}{HTML}{E8E4D7}

\newtcolorbox{conceptbox}[2]{%
  enhanced,
  breakable,
  colback=#1!6,
  colframe=#1!75!black,
  colbacktitle=#1!16,
  coltitle=black,
  fonttitle=\bfseries,
  title={#2},
  boxrule=0.55pt,
  arc=1.2mm,
  left=1.2mm,
  right=1.2mm,
  top=0.8mm,
  bottom=0.8mm,
  before skip=6pt,
  after skip=6pt
}

\newcounter{surveybox}

\newcommand{\glossaryitem}[2]{%
  \par\noindent
  \begingroup
  \parindent=0pt
  \parshape=2
    1.2em \dimexpr\linewidth-1.2em\relax
    0pt \linewidth
  \textbf{#1} #2\par
  \endgroup
  \vspace{0.35em}%
}

\definecolor{scopefill}{HTML}{EEF3FB}
\definecolor{scopeframe}{HTML}{4D63C7}
\definecolor{scopehead}{HTML}{DCE4FA}

\definecolor{contribfill}{HTML}{FBF1E2}
\definecolor{contribframe}{HTML}{E28B20}
\definecolor{contribhead}{HTML}{F6E1BF}

\definecolor{structfill}{HTML}{ECECF7}
\definecolor{structframe}{HTML}{4A50B7}
\definecolor{structhead}{HTML}{D9DCEE}

\tcbset{
  surveycallout/.style={
    enhanced,
    breakable,
    boxrule=0.6pt,
    arc=2mm,
    left=2.0mm,
    right=2.0mm,
    top=1.0mm,
    bottom=1.0mm,
    coltitle=black,
    fonttitle=\bfseries,
    titlerule=0pt,
    toptitle=0.8mm,
    bottomtitle=0.8mm,
    before skip=8pt,
    after skip=8pt
  }
}

\newtcolorbox{surveyScopeBox}{
  surveycallout,
  title={Survey Scope},
  colback=scopefill,
  colframe=scopeframe,
  colbacktitle=scopehead
}

\newtcolorbox{contributionBox}{
  surveycallout,
  title={Contributions},
  colback=contribfill,
  colframe=contribframe,
  colbacktitle=contribhead
}

\newtcolorbox{surveyStructureBox}{
  surveycallout,
  title={Survey Structure},
  colback=structfill,
  colframe=structframe,
  colbacktitle=structhead
}

\usepackage{color}

\newcolumntype{L}[1]{>{\raggedright\arraybackslash}p{#1}}
\newcommand{\approwspace}{\addlinespace[4pt]}

\newcommand{\Template}{\bar{\mathcal{G}}}
\newcommand{\Realized}{\mathcal{G}^{\mathrm{run}}}
\newcommand{\Trace}{\tau}

\newcommand{\Yes}{\ensuremath{\checkmark}}
\newcommand{\No}{\ensuremath{\times}}
\newcommand{\Partial}{\ensuremath{\triangle}}

\theoremstyle{definition}
\newtheorem{definition}{Definition}

\title{From Static Templates to Dynamic Runtime Graphs: \\A Survey of Workflow Optimization for LLM Agents}

\author{\name Ling Yue \email yuel2@rpi.edu \\
\addr Department of Computer Science, Rensselaer Polytechnic Institute, Troy, NY 12180 USA
\AND
\name Kushal Raj Bhandari \email bhandk@rpi.edu \\
\addr Department of Computer Science, Rensselaer Polytechnic Institute, Troy, NY 12180 USA
\AND
\name Ching-Yun Ko \email cyko@ibm.com \\
\addr IBM Research, Yorktown Heights, NY 10598 USA
\AND
\name Dhaval Patel \email pateldha@us.ibm.com \\
\addr IBM Research, Yorktown Heights, NY 10598 USA
\AND
\name Shuxin Lin \email shuxin.lin@ibm.com \\
\addr IBM Research, Yorktown Heights, NY 10598 USA
\AND 
\name Nianjun Zhou \email jzhou@us.ibm.com \\
\addr IBM Research, Yorktown Heights, NY 10598 USA
\AND
\name Jianxi Gao \email gaoj8@rpi.edu \\
\addr {Department of Computer Science, Rensselaer Polytechnic Institute, Troy, NY 12180 USA}
\AND
\name Pin-Yu Chen\footnotemark[1] \email pin-yu.chen@ibm.com \\
\addr IBM Research, Yorktown Heights, NY 10598 USA
\AND
\name Shaowu Pan\thanks{Corresponding authors.} \email pans2@rpi.edu \\
\addr Department of Mechanical, Aerospace, and Nuclear Engineering, Rensselaer Polytechnic Institute, Troy, NY 12180 USA
}

\begin{document}
\maketitle

% ============================================================
\begin{abstract}
Large language model (LLM)-based systems are becoming increasingly popular for solving tasks by constructing executable workflows that interleave LLM calls, information retrieval, tool use, code execution, memory updates, and verification. 
This survey reviews recent methods for designing and optimizing such workflows, which we treat as agentic computation graphs (ACGs). 
We organize the literature based on when workflow structure is determined, where `structure' refers to which components or agents are present, how they depend on each other, and how information flows between them.
This lens distinguishes static methods, which fix a reusable workflow scaffold before deployment, from dynamic methods, which select, generate, or revise the workflow for a particular run before or during execution.
We further organize prior work along three dimensions: when structure is determined, what part of the workflow is optimized, and which evaluation signals guide optimization (e.g., task metrics, verifier signals, preferences, or trace-derived feedback).
We also distinguish reusable workflow templates, run-specific realized graphs, and execution traces, separating reusable design choices from the structures actually deployed in a given run and from realized runtime behavior. 
Finally, we outline a structure-aware evaluation perspective that complements downstream task metrics with graph-level properties, execution cost, robustness, and structural variation across inputs.
Our goal is to provide a clear vocabulary, a unified framework for positioning new methods, a more comparable view of existing body of literature, and a more reproducible evaluation standard for future work in workflow optimizations for LLM agents.
\begingroup
\renewcommand\thefootnote{}
\footnotetext{\faGithub\: \url{https://github.com/IBM/awesome-agentic-workflow-optimization}.}
\addtocounter{footnote}{-1}
\endgroup
\end{abstract}

% ============================================================
% Auto-generated table of contents
% ============================================================
\phantomsection
\pdfbookmark[1]{Contents}{toc}
\setcounter{tocdepth}{2}
\tableofcontents

\begin{surveyStructureBox}
This survey moves from concepts to methods, evaluation, and future directions.

\begin{itemize}[leftmargin=*,itemsep=0pt, topsep=0pt]
    \item \textbf{Sec.~\ref{sec:concept}} introduces the conceptual framework and taxonomy.
    
    \item \textbf{Secs.~\ref{sec:static} and \ref{sec:dynamic}} review static and dynamic workflow optimization methods.
    
    \item \textbf{Sec.~\ref{sec:signals}} organizes these methods by feedback signals and update mechanisms.
    
    \item \textbf{Secs.~\ref{sec:eval} and \ref{sec:synthesis}} discuss evaluation, reporting, and design trade-offs.
    
    \item \textbf{Sec.~\ref{sec:future}} highlights open problems and future directions.
\end{itemize}
\end{surveyStructureBox}

\begin{figure*}[htbp]
\refstepcounter{surveybox}\label{box:glossary}
\centering
\begin{tcolorbox}[
  enhanced,
  width=\textwidth,
  colback=glossbg,
  colframe=black!80,
  boxrule=0.45pt,
  arc=0pt,
  left=3.6mm,
  right=3.6mm,
  top=2.6mm,
  bottom=3.0mm
]
{\Large\bfseries Glossary}\par
\vspace{2.4mm}
\small
\begin{minipage}[t]{0.485\textwidth}
\glossaryitem{Agentic computation graph (ACG)}{our unifying abstraction for an executable LLM-centered workflow. Nodes perform atomic actions such as LLM calls, information retrieval, tool use, validation, or message passing. Edges encode control, data, or communication dependencies.}
\glossaryitem{ACG template ($\Template$)}{reusable executable specification $(\mathcal{V},\mathcal{E},\Phi,\Sigma,\mathcal{A})$.}
\glossaryitem{Realized graph ($\Realized$)}{workflow structure actually used for a particular run.}
\glossaryitem{Execution trace ($\Trace$)}{sequence of states, actions, observations, and costs produced by executing $\Realized$.}
\glossaryitem{Workflow}{an executable organization of multiple steps for solving a task.}
\glossaryitem{Workflow structure}{which components or agents are present, how they depend on one another, and how information flows between them.}
\glossaryitem{Topology}{the communication pattern among agents or modules.}
\end{minipage}\hfill
\begin{minipage}[t]{0.485\textwidth}
\glossaryitem{Graph determination time (GDT)}{when structure is decided: \emph{offline}, \emph{pre-execution}, or \emph{in-execution}.}
\glossaryitem{Graph plasticity mode (GPM)}{how much structure can change at inference time: \emph{none}, \emph{select}, \emph{generate}, or \emph{edit}.}
\glossaryitem{Node parameters ($\Phi$)}{prompts, role descriptions, tool schemas, model choices, decoding settings, verifier settings, and memory policies.}
\glossaryitem{Structure variables}{topology, scheduling or routing policy, activation choices, and admissible edit actions.}
\glossaryitem{Scope tags}{\textsc{core}: directly optimizes reusable templates or executable realized graphs. \textsc{adjacent}: changes the effective workflow through routing, team selection, pruning, or communication sparsification. \textsc{background}: provides frameworks, datasets, or benchmarks that shape workflow optimization or evaluation.}
\end{minipage}
\end{tcolorbox}
\end{figure*}

\clearpage

% ============================================================
\section{Introduction}
\label{sec:intro}

\begin{figure*}[t]
\centering
\includegraphics[width=\textwidth]{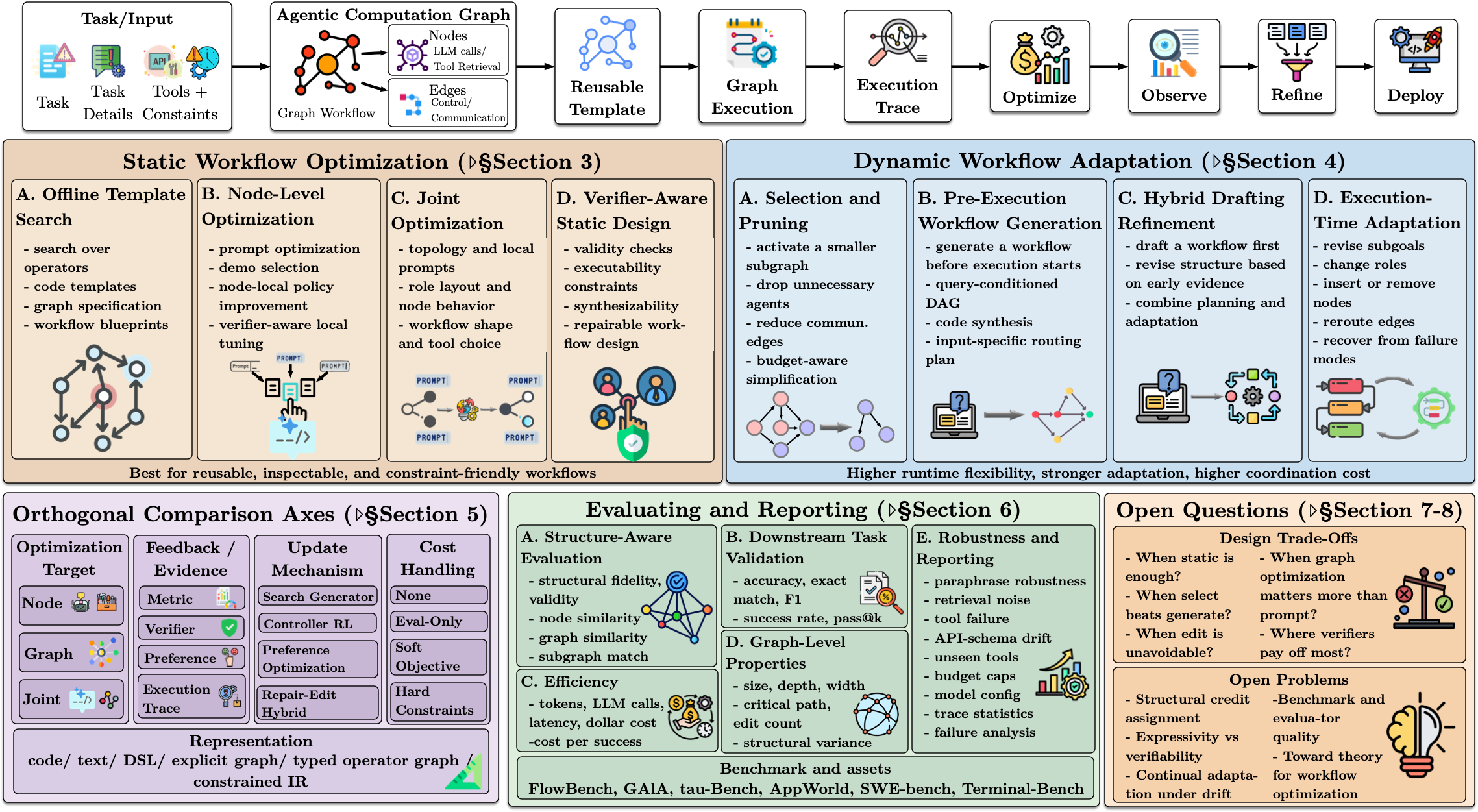}
\caption{An overview of workflow optimization for LLM Agents.}
\label{fig:survey-overview}
\end{figure*}

Large language model (LLM) systems are evolving beyond simple chatbots that generate responses to single prompts. Instead, they are increasingly being integrated into executable workflows that coordinate multiple actions over time. By workflow, we mean an executable organization of multiple steps, such as LLM calls, tool use, information retrieval, code execution, memory updates, and verification, to accomplish a task. In practice, a system may need to decompose a task, call tools, retrieve documents, execute code, update memory, verify intermediate results, and recover from failures. For example, a coding assistant may retrieve relevant files, propose edits, run tests, and use a verifier to decide whether to revise or stop. In multi-agent systems (MAS), these actions may be distributed across multiple specialized agents that communicate over a defined communication pattern, which specifies how agents are connected and how messages flow between them.
What matters in practice is not only the quality of each individual model call, but also the overall workflow structure that determines what is called, when it is called, and how information flows between calls. Here, the workflow structure refers to the components or agents present, how they depend on each other, and how information flows between them. Once an agentic system is represented as a graph, one can reason about topology, communication density, scheduling, verification placement, and cost. These design choices often affect both effectiveness and efficiency \citep{zhangAFlowAutomatingAgentic,zhouMultiAgentDesignOptimizing2025,liAdaptiveGraphPruning2025}. A weak scaffold can sometimes be rescued by better prompts, but it can also be improved by adding a verifier, such as a unit-test stage or a schema checker, pruning redundant communication, changing a manager--worker hierarchy, or replacing a fixed, one-size-fits-all pipeline with run-specific generation. However, improvements in agent capability often come with hidden structural costs, such as excessive depth, fragile control flow, and high communication overhead.

% Terminology is not uniform across the literature. 
In this survey, we use \emph{workflow} in a broad structural sense. Under this view, both fixed pipelines and more autonomous agentic systems can be studied as executable organizations of nodes, dependencies, and control decisions. The difference is how much of the structure is fixed before deployment versus determined for a particular run or revised during execution. We use the term \emph{agentic computation graph} (ACG) as a unifying abstraction for executable LLM-centered workflows. The term brings together work scattered across different names in the literature: workflows, pipelines, orchestration graphs, communication graphs, plans, and code-defined agent systems. Our goal is not to impose new terminology for its own sake, but to make structure itself the primary object of comparison.

A growing body of work now treats workflow design as an optimization problem. Some works search for reusable templates offline \citep{zhangAFlowAutomatingAgentic,huAutomatedDesignAgentic2025,zhouMultiAgentDesignOptimizing2025}. Others optimize prompts, demonstrations, or collaboration behavior within a fixed scaffold \citep{khattabDSPyCompilingDeclarative2023,yangLargeLanguageModels2024,guoEvoPromptConnectingLLMs2025,zehleCAPOCostAwarePrompt2025,agrawalGEPAReflectivePrompt2025,chenOptimaOptimizingEffectiveness2025}. A third group generates, selects, or edits the workflow used for a particular run before or during execution \citep{liAssembleYourCrew2025,zhangGDesignerArchitectingMultiagent2025,liAdaptiveGraphPruning2025,gaoFlowReasonerReinforcingQueryLevel2025,Wang2026AgentConductorTE,wangMetaGenSelfEvolvingRoles2026}. This distinction matters because these methods optimize different artifacts: reusable templates, local behavior inside a fixed scaffold, or the realized workflow structure used for a given run. Across these lines of work, the central question is no longer only what capability an agent has, but also what workflow structure should be used, when that structure should be determined, and how it should be optimized under quality--cost trade-offs.

To the best of our knowledge, existing surveys focused on workflow and infrastructure are limited to the ecosystem of agent systems, engineering abstractions, and orchestration frameworks \citep{yuSurveyAgentWorkflow2025,liSurveyLLMbasedMultiagent2024}. Other surveys focused on the workflow planning phase emphasize decomposition, reflection, memory, and external modules as ingredients in agent planning \citep{huangUnderstandingPlanningLLM2024}. Tool-learning surveys focus on retrieving, selecting, and invoking tools \citep{xuLLMBasedAgentsTool2025}. Multi-agent surveys organize the literature by collaboration mechanisms, communication protocols, and application domains \citep{chenSurveyLLMbasedMultiAgent2025,tranMultiAgentCollaborationMechanisms2025,zhangSurveyMultiAIAgent2025,buyya2026agentic}. Broader optimization surveys cover many ways to improve LLM agents, often through the lens of parameter-driven versus parameter-free methods \citep{duSurveyOptimizationLarge2026,yue2026building}. While these surveys provide important foundations, the design of the workflow structure is usually taken as given rather than treated as the primary optimization target. In most papers, graph construction is implied as code, a communication pattern, or a planner--executor loop, rather than being treated as a first-class optimization object to be searched, generated, edited, or evaluated. 

Existing surveys mostly cover adjacent slices of the agent literature rather than workflow optimization itself. Table~\ref{tab:survey-positioning} positions our survey within that broader landscape and clarifies the specific gap it fills. To make the boundary of this survey explicit, we summarize the scope below.

\begin{surveyScopeBox}
This survey studies workflow optimization for LLM-centered agentic systems, where workflow structure determines how LLM calls, retrieval, tool use, code execution, memory, and verification are composed.

\begin{itemize}[leftmargin=*]
    \item \textbf{Static workflow optimization:} improves a reusable template offline, including scaffold search, topology design, and node-level optimization inside a fixed structure.
    \item \textbf{Dynamic workflow optimization:} determines part of the realized graph at inference time through subgraph selection, pre-execution generation, or in-execution editing.
\end{itemize}

Our primary scope is methods that treat workflow structure itself as the optimization target. We exclude planning-only or tool-use-only papers unless they directly alter workflow structure or provide workflow-relevant evaluation assets. Throughout the paper, we use three scope tags: \textsc{core} for methods that directly optimize reusable templates or executable realized graphs, \textsc{adjacent} for routing, team selection, and pruning methods that change the effective workflow, and \textsc{background} for frameworks, datasets, and benchmarks that materially shape workflow optimization or evaluation.
\end{surveyScopeBox}

To ground this survey in a focused yet broad evidence base, we compiled an inventory of 77 in-scope works, including 39 \textsc{core} papers, 7 \textsc{adjacent} papers, and 31 \textsc{background} resources.
Separately, because evaluation is itself a major part of this literature, we organize 27 workflow-relevant evaluation assets, including 20 benchmark or environment papers and 7 dataset, training-corpus, or validator resources.
The surveyed materials span archival arXiv preprints, peer-reviewed conference and workshop papers from major ML and NLP venues, open-source frameworks, and benchmark or dataset resources that define much of the current state of practice.
A work is included if at least one of the following holds: it optimizes a reusable workflow template, generates, selects, or edits a realized graph for a specific run, studies communication topology or routing structure as an optimization variable, provides infrastructure that strongly shapes this design space, or contributes a benchmark or dataset used specifically to assess workflow generation or agent execution.
The goal is a structured, in-scope synthesis of a fast-moving literature.

\begin{contributionBox}
This survey makes the following contributions:

\begin{itemize}[leftmargin=*]
    \item \textbf{Workflow-centered formulation:} We introduce a workflow-centered view of LLM agent systems as agentic computation graphs (ACGs) and distinguish reusable templates, run-specific realized graphs, and execution traces. This separation clarifies whether a method optimizes reusable design, per-run structure, or realized runtime behavior.
    
    \item \textbf{Taxonomy of structure determination:} We propose a taxonomy organized by when workflow structure is determined, refined by graph determination time (GDT) and graph plasticity mode (GPM) to compare fixed-template optimization, pre-execution generation or selection, and in-execution editing under a common lens. This taxonomy helps resolve gray cases such as offline-trained workflow generators and runtime subgraph selection.
    
    \item \textbf{Cross-cutting synthesis:} We synthesize the literature along three orthogonal axes---optimization target (node, graph, or joint), evidence source (metric, verifier, preference, or trace), and update mechanism---to explain what part of the workflow each method changes, what evidence authorizes those changes, and how quality--cost trade-offs are managed.
    
    \item \textbf{Evaluation protocol:} We organize workflow-relevant evaluation assets and propose a minimum reporting protocol that separates downstream task performance from graph properties, execution cost, robustness, and structural variation across inputs, thereby making workflow evaluation more comparable and reproducible.
\end{itemize}
\end{contributionBox}

\begin{table}[htbp]
\centering
\footnotesize
\setlength{\tabcolsep}{3pt}
\renewcommand{\arraystretch}{1.08}
\caption{Positioning this survey in the broader survey landscape. Most prior surveys center on adjacent topics such as planning, tool learning, collaboration, architecture, or broad agent optimization, whereas this survey centers workflow structure itself.}
\label{tab:survey-positioning}
\begin{tabularx}{\linewidth}{@{}L{0.23\linewidth}L{0.20\linewidth}L{0.18\linewidth}>{\raggedright\arraybackslash}X@{}}
\toprule
Survey family & Central lens & How workflow appears & Relation to this survey \\
\midrule

Workflow / infrastructure\newline
\citep{yuSurveyAgentWorkflow2025,liSurveyLLMbasedMultiagent2024}
& workflow systems, orchestration abstractions, and MAS system organization
& workflow as system scaffold, lifecycle, or engineering abstraction
& adjacent but not centered on structure search, per-run graph construction, or workflow-specific evaluation \\
\approwspace

Planning\newline
\citep{huangUnderstandingPlanningLLM2024}
& decomposition, plan selection, reflection, memory, and external planning modules
& workflow as one internal phase or capability of an agent
& covers a crucial ingredient of agents, but not workflow optimization as a broader graph-design problem \\
\approwspace

Tool learning\newline
\citep{xuLLMBasedAgentsTool2025}
& tool retrieval, tool planning, tool invocation, and tool-use evaluation
& workflow as a tool-use pipeline and environment interface
& centers tool competence rather than topology, routing, or graph generation/editing \\
\approwspace

MAS collaboration / applications\newline
\citep{chenSurveyLLMbasedMultiAgent2025,tranMultiAgentCollaborationMechanisms2025,zhangSurveyMultiAIAgent2025}
& collaboration mechanisms, communication protocols, and application domains
& workflow as a communication pattern or application-specific coordination scheme
& studies collaborative behavior and applications, not reusable-template or realized-graph design as the main object \\
\approwspace

Broad agent optimization\newline
\citep{duSurveyOptimizationLarge2026}
& parameter-driven and parameter-free optimization of agents overall
& workflow as one optimization lever among prompts, training, retrieval, memory, and other factors
& does not isolate workflow structure as the primary object of comparison \\
\approwspace

Agent architecture / evaluation\newline
\citep{buyya2026agentic}
& architectures, taxonomies, and evaluation of agentic AI systems
& workflow as one architectural manifestation within a larger agent taxonomy
& provides a broad architecture-and-evaluation view, but not a workflow-optimization-specific synthesis \\
\approwspace

\midrule
Workflow Optimization (ours)
& workflow optimization for LLM agents
& workflow structure as the primary optimization object
& unifies static template optimization, per-run graph generation/selection/editing, template--realized graph--trace distinctions, and structure-aware evaluation/reporting \\
\bottomrule
\end{tabularx}
\end{table}

\section{Conceptual Framework and Taxonomy}
\label{sec:concept}

This section introduces the distinctions that organize the rest of the survey: reusable templates versus realized graphs, node-level versus graph-level optimization, and static versus dynamic structure determination. Box~\ref{box:glossary} provides a compact reference for the main terms and notation used below.

\subsection{Agentic computation graphs as executable workflows}

An \emph{agentic computation graph} (ACG) is our unifying abstraction for an executable LLM-centered workflow. Nodes perform atomic actions such as LLM calls, information retrieval, tool use, validation, or message passing. Edges encode control, data, or communication dependencies.

Typical nodes include LLM calls, information retrieval, code execution, database access, tool invocation, validation, memory updates, and message-passing steps between specialized agents. In multi-agent systems (MAS), the same abstraction covers role allocation and communication topology: agents correspond to LLM-driven nodes with distinct prompts, tools, or models, and their messages appear as graph edges. 

A convenient way to describe an LLM-centered node is to use the tuple 
$\langle$ \text{Instruction},\ \text{Context},\ \text{Tools},\ \text{Model/Decoding} $\rangle$,
which is general enough to cover both single-agent modular pipelines and multi-agent systems with heterogeneous capabilities. Beyond nodes and edges, many workflows also include a scheduler or router that decides which node executes next, which nodes can run in parallel, when to terminate, and whether replanning is allowed. 

Different papers realize this abstraction differently. Some use code-defined workflows, where graph structure is implicit in control flow. Others use a domain-specific language (DSL), JSON, YAML, or a plain-text workflow specification, which is often easier to generate or edit but varies in how easily it can be validated. Still others use explicit graph intermediate representations (IRs) with typed operators or constrained schemas, especially when executability and validity are central concerns. Representation matters because it shapes what can be searched, verified, or edited. 
\subsection{Template, realized graph, and trace}

We distinguish three related but different objects throughout the survey: a reusable workflow template, the realized graph for a particular run, and the execution trace produced by running that graph.

\begin{definition}[ACG template]
An \emph{ACG template} is a reusable executable specification
\[
\Template = (\mathcal{V}, \mathcal{E}, \Phi, \Sigma, \mathcal{A}),
\]
where $\mathcal{V}$ is the set of nodes, $\mathcal{E}$ is the set of directed edges, $\Phi=\{\phi_v\}_{v\in\mathcal{V}}$ contains node parameters such as prompts, tool schemas, model choices, or verifier settings, $\Sigma$ is a scheduling or routing policy, and $\mathcal{A}$ is the set of admissible activation or edit actions.
\end{definition}

The template is the reusable design object. It specifies the structural and parametric space available to the system before a concrete input is observed.

\begin{definition}[Realized graph]
Given an input $x$ and the evolving runtime state, a \emph{realized graph} $\Realized$ is the workflow structure actually used for a particular run. It may coincide with the reusable template, or it may be obtained by selecting a subgraph, instantiating optional nodes, or applying allowed edits before or during execution.
\end{definition}

The realized graph is therefore run-specific. It captures the structure that is actually deployed for one run, rather than the structure merely available in principle. Every executed workflow has a realized graph; a method is dynamic not because such a graph exists, but because some of its structure is chosen or revised at inference time rather than being fully fixed in the reusable template.

\begin{definition}[Execution trace]
Executing $\Realized$ yields an \emph{execution trace}
\[
\Trace = \{(s_t,a_t,o_t,c_t)\}_{t=1}^{T},
\]
where $s_t$ is the system or environment state, $a_t$ is the action taken, $o_t$ is the resulting observation, and $c_t$ is execution cost such as token usage, tool calls, latency, or monetary expense.
\end{definition}

These definitions are intentionally lightweight. The template captures what a method makes reusable. The realized graph captures what structure is actually deployed for a particular run. The trace records what happened during execution, including tool failures, retries, verifier outputs, and cost accumulation. Many recent papers differ precisely in which of these three objects is optimized, generated, or reused.

Two simple examples make the distinction concrete. In a fixed planner--retriever--executor--verifier pipeline, the pipeline definition is the template, and the workflow structure that is actually traversed for one question---often the full pipeline, but sometimes a pre-authored branch-conditioned substructure---is the realized graph. The resulting retrieval calls, code executions, verifier decisions, and retries constitute the trace. By contrast, in a query-conditioned multi-agent workflow generator, the reusable object may specify only the operator vocabulary or graph-generation policy; each query then induces a different realized graph, and the trace records the messages, tool calls, failures, and edits that occur when that generated graph runs.

\subsection{A quality--cost view of workflow optimization}

Across a wide range of methods, workflow optimization can be viewed as balancing task quality against execution cost. Let $R(\Trace;x)$ denote a task-quality score, such as success, accuracy, pass@k, or an application-specific measure, and let $C(\Trace)$ denote execution cost. A convenient formulation is
\begin{equation}
\max \;\; \mathbb{E}_{x\sim\mathcal{D}}\left[\mathbb{E}_{\Realized \mid x}\left[\mathbb{E}_{\Trace \mid \Realized, x}\left[R(\Trace;x)-\lambda C(\Trace)\right]\right]\right],
\label{eq:acg_objective}
\end{equation}
where the inner expectation reflects execution stochasticity conditional on a realized graph, the middle expectation matters when the method generates or selects a run-specific graph, and $\lambda$ controls the quality--cost trade-off. This expression is schematic. For in-execution editing, the realized graph can be treated as part of the evolving system state, and edit actions appear inside the trace rather than being decided entirely upfront.

This formulation also clarifies three recurring optimization targets. In \emph{node-level optimization}, the high-level scaffold is fixed and local parameters in $\Phi$---such as prompts, tools, models, or verifier policies---are improved. In \emph{graph-level optimization}, structural variables such as $\mathcal{E}$, $\Sigma$, and $\mathcal{A}$ are updated, changing topology, communication structure, scheduling, branching logic, or the admissible edit space. In \emph{joint optimization}, both are updated together, either simultaneously or in alternating stages. This distinction helps explain why two methods that both improve final accuracy may in fact be optimizing very different parts of the workflow.

\begin{table}[!t]
\centering
\footnotesize
\caption{Representative \textsc{core} static workflow-optimization methods, summarized with a unified comparison card. When a paper uses multiple evidence types or update mechanisms, columns report the dominant workflow-relevant entry in each field.}
\label{tab:static-papers}
\begin{tabular}{@{}L{0.18\linewidth}L{0.12\linewidth}L{0.08\linewidth}L{0.14\linewidth}L{0.15\linewidth}L{0.15\linewidth}L{0.14\linewidth}@{}}
\toprule
Method & Setting (GDT/GPM) & Level & Representation & Feedback / evidence & Update mechanism & Cost handling \\
\midrule
AFlow \citep{zhangAFlowAutomatingAgentic} & offline / none & graph & typed operator graph & metric: task score & search: MCTS & soft objective (\$) \\
\approwspace
ADAS \citep{huAutomatedDesignAgentic2025} & offline / none & joint & runnable code & metric: task score & search: archive meta-search & none \\
\approwspace
A$^2$Flow \citep{zhaoA2FlowAutomatingAgentic2025} & offline / none & graph & abstract operator graph & supervision: demos + execution & hybrid: operator learning + search & none \\
\approwspace
Multi-Agent Design \citep{zhouMultiAgentDesignOptimizing2025} & offline / none & joint & topology + prompts & metric: task score & hybrid: staged alternation & none \\
\approwspace
Learning Multi-Agent Communication from Graph Modeling Perspective \citep{huLearningMultiAgentCommunication2024} & offline / none & joint & differentiable communication graph & proxy: training objective & continuous relaxation & none \\
\approwspace
Optima \citep{chenOptimaOptimizingEffectiveness2025} & offline / none & node & fixed scaffold + trajectories & metric: quality--efficiency reward & hybrid: generate-rank-select-train & soft objective (efficiency) \\
\approwspace
SEW \citep{liuSEWSelfEvolvingAgentic2025} & offline / none & graph & text workflow spec. & metric: pass rate & search: self-evolution & none \\
\approwspace
VFlow \citep{weiVFlowDiscoveringOptimal2025} & offline / none & graph & domain workflow graph & verifier: multi-level checks & hybrid: MCTS + cooperative evolution & soft objective (resource) \\
\approwspace
MermaidFlow \citep{zhengMermaidFlowRedefiningAgentic2025} & offline / none & graph & constrained Mermaid IR & verifier: validity + execution & search: constrained evolution & none \\
\approwspace
Maestro \citep{wang2025maestro} & offline / none & joint & typed stochastic graph & trace: reflective text + score & hybrid: alternating graph/config updates & soft objective (budget) \\
\approwspace
Evolutionary Generation of Multi-Agent Systems \citep{huEvolutionaryGenerationMultiAgent2026} & offline / none & graph & evolvable MAS genotype & metric: execution fitness & search: evolutionary & none \\
\bottomrule
\end{tabular}
\end{table}

\subsection{Structure determination}

Our main organizing principle is \emph{when workflow structure is determined}. We distinguish between static structure determination and dynamic structure determination, and we use two lightweight descriptors to clarify gray cases.

\subsubsection{Static structure determination}
A method is \emph{static} if its deployed structure is a reusable template whose structural degrees of freedom are fixed after training or search. The template can still contain conditional execution, loops, or stochasticity, but those behaviors are already encoded in the reusable scaffold. A pre-authored branch such as ``if retrieval fails, retry once'' remains static because the branching logic is fixed in the template.

\subsubsection{Dynamic structure determination}
A method is \emph{dynamic} if some part of the realized graph is constructed, selected, or edited at inference time. This may happen once before execution, or repeatedly during execution in response to observations, failures, or verifier signals. Adding a new verifier, spawning a new agent, or rewiring communication only after observing execution feedback are therefore dynamic changes.

To compare gray cases, we use two lightweight descriptors. \emph{Graph determination time} (GDT) records when the realized structure is decided: \emph{offline} if the reusable template is optimized before deployment, \emph{pre-execution} if a run-specific graph is generated once before execution, and \emph{in-execution} if structure is revised during execution. \emph{Graph plasticity mode} (GPM) records how structure may vary at inference time: \emph{none} if the structure is fixed, \emph{select} if the method activates or prunes parts of a fixed super-graph, \emph{generate} if it constructs a run-specific graph before execution, and \emph{edit} if it adds, removes, rewires, or rewrites structure during execution.

These descriptors are deliberately lightweight. Their purpose is to clarify common ambiguities rather than to impose a rigid ontology. For example, a workflow generator trained offline but used to emit a new plan for each input is dynamic in our taxonomy, because the realized graph is determined pre-execution at inference time. Likewise, a fixed super-graph with inference-time pruning is meaningfully dynamic even if it never creates a wholly new workflow from scratch.

\subsection{Comparison card}
For comparability, the main-text tables summarize papers using a compact classification card. The card records the structural setting (static or dynamic, together with GDT and GPM), optimized level (node, graph, or joint), representation, dominant feedback/evidence used to accept or revise structures, dominant update mechanism, and cost handling. We intentionally omit a free-form application-scenario column from the core card, because scenario descriptions are useful but are not part of the controlled comparison schema; application context is instead discussed in the surrounding text. The main-text comparison tables use this card consistently so that methods can be compared along stable dimensions rather than paper-specific descriptions. When a paper uses multiple evidence types or update mechanisms, we report the dominant workflow-relevant entry in each field, with finer distinctions discussed in the surrounding text.

% ============================================================
\section{Static Optimization of Agent Workflows}
\label{sec:static}

Static methods optimize a reusable template or a fixed collaboration scaffold before deployment. Their practical appeal is clear: they are easier to inspect, constrain, ablate, and benchmark under stable budgets. The main limitation is equally clear: once a template is frozen, distribution shift, tool drift, or unanticipated branching can expose brittle structural assumptions. Table~\ref{tab:static-papers} summarizes representative \textsc{core} static methods using the same comparison card later applied to dynamic ones.

\subsection{Offline template search over constrained design spaces}
A central line of work treats the workflow as a discrete design object and searches for a reusable template by repeated execution and evaluation. AFlow~\citep{zhangAFlowAutomatingAgentic} searches over typed operator graphs using Monte Carlo Tree Search (MCTS), combining LLM-guided expansion with executable evaluation and explicit dollar cost. Automated Design of Agentic Systems (ADAS)~\citep{huAutomatedDesignAgentic2025} instead searches in code space: a meta-agent proposes runnable agentic systems, evaluates them, archives strong designs, and iteratively improves them. Evolutionary Generation of Multi-Agent Systems~\citep{huEvolutionaryGenerationMultiAgent2026} adopts a more classical population-based view and treats multi-agent system design as an evolvable genotype of roles, topology, and protocol. In a domain-specific setting, VFlow~\citep{weiVFlowDiscoveringOptimal2025} combines cooperative evolution and Monte Carlo Tree Search with strong hardware verifiers to discover Verilog-generation workflows under functional and resource objectives.

These methods differ in representation, but they share three assumptions. First, there must be an executable search space, whether defined by typed operators, code templates, or structured workflow languages. Second, evaluation must be reliable enough to discriminate candidates. Third, the search space must embody a useful inductive bias: if candidate workflows are mostly invalid or semantically incoherent, black-box search quickly becomes prohibitively expensive. This is precisely why typed operators, code scaffolds, and constrained graph languages are so important in practice.

A notable variation is to learn the operator library itself instead of assuming it is fixed. A$^2$Flow~\citep{zhaoA2FlowAutomatingAgentic2025} extracts abstract operators from demonstrations, clusters them into reusable patterns, and then refines them for later workflow search. This reduces dependence on hand-designed primitives and makes explicit a design choice that many search papers leave implicit: the quality of the operator vocabulary often matters as much as the search algorithm. SEW~\citep{liuSEWSelfEvolvingAgentic2025} sits at the opposite end of the representation spectrum. It encodes candidate workflows directly as text and iteratively self-evolves them using benchmark pass rates in code-generation settings, showing that reusable workflow optimization need not always rely on a typed graph IR if the task feedback is strong enough. More generally, this line of work suggests a recurring theme of the survey: search quality is often limited less by the nominal optimizer than by the representation and evaluator it is allowed to use.

\subsection{Node-level optimization inside fixed scaffolds}
A fast path to improvement is to keep the scaffold fixed and optimize local components. DSPy~\citep{khattabDSPyCompilingDeclarative2023} is the clearest programmatic example: an LLM pipeline is treated as a composition of modules, and a compiler synthesizes prompts or demonstrations that optimize a user-specified metric on examples. The underlying graph is fixed, but the local node policies are no longer hand-written strings.

Black-box prompt optimizers such as Large Language Models as Optimizers (OPRO)~\citep{yangLargeLanguageModels2024}, EvoPrompt~\citep{guoEvoPromptConnectingLLMs2025}, CAPO~\citep{zehleCAPOCostAwarePrompt2025}, and GEPA~\citep{agrawalGEPAReflectivePrompt2025} fit naturally into this picture when prompts are attached to workflow nodes rather than viewed as isolated artifacts. OPRO~\citep{yangLargeLanguageModels2024} uses an LLM as the optimizer over a textual history of candidates and scores. EvoPrompt~\citep{guoEvoPromptConnectingLLMs2025} turns prompt search into evolutionary optimization. CAPO~\citep{zehleCAPOCostAwarePrompt2025} makes the same family explicitly cost-aware through racing and prompt-length penalties. GEPA~\citep{agrawalGEPAReflectivePrompt2025} uses trajectory-level reflection and Pareto-style retention of prompt edits, showing that textual feedback can outperform pure scalar-reward optimization when failures are semantically structured. Because these methods optimize workflow-local instructions rather than global topology, we treat them as node-level optimizers; Appendix Table~\ref{tab:prompt-papers-appendix} summarizes them in a compact comparison matrix.

The value of node-level optimization is practical: it usually requires fewer assumptions than graph search and can yield large gains quickly. The main limitation is interpretive rather than purely empirical. Better prompts can compensate for weak topology, making a poor scaffold look competitive while increasing cost and reducing robustness. Static node tuning is therefore often a necessary baseline, but rarely a sufficient explanation for why a workflow works.

A closely related but more training-oriented line optimizes collaboration behavior within a largely fixed multi-agent scaffold. Optima~\citep{chenOptimaOptimizingEffectiveness2025} generates, ranks, selects, and trains on multi-agent interaction trajectories to improve the effectiveness--efficiency trade-off of collaboration itself. In our taxonomy, this remains static at deployment time if the collaboration scaffold is fixed, even though the underlying models are updated.

{\scriptsize
\begin{longtable}{@{}L{0.19\linewidth}L{0.12\linewidth}L{0.08\linewidth}L{0.14\linewidth}L{0.15\linewidth}L{0.15\linewidth}L{0.13\linewidth}@{}}
\caption{Representative \textsc{core} dynamic workflow-optimization methods, summarized with the same comparison card as Table~\ref{tab:static-papers}. When a paper uses multiple evidence types or update mechanisms, columns report the dominant workflow-relevant entry in each field.}\label{tab:dynamic-papers}\\
\toprule
Method & Setting (GDT/GPM) & Level & Representation & Feedback / evidence & Update mechanism & Cost handling \\
\midrule
\endfirsthead

\toprule
Method & Setting (GDT/GPM) & Level & Representation & Feedback / evidence & Update mechanism & Cost handling \\
\midrule
\endhead

\multicolumn{7}{@{}l}{\textit{Pre-execution generation or selection}}\\
\midrule
Difficulty-Aware Agentic Orchestration \citep{su2025difficulty} & pre-exec / select & joint & modular operator workflow & proxy: difficulty estimate & controller: router + allocator & soft objective (cost) \\
\approwspace
Assemble Your Crew \citep{liAssembleYourCrew2025} & pre-exec / generate & graph & query-conditioned DAG & supervision: task labels & generator: autoregressive DAG & none \\
\approwspace
G-Designer \citep{zhangGDesignerArchitectingMultiagent2025} & pre-exec / generate & graph & generated communication graph & metric: quality score & generator: VGAE & soft objective (cost) \\
\approwspace
Dynamic Generation of Multi-LLM Agents Communication Topologies \citep{jiang2025dynamic} & pre-exec / generate & graph & diffusion-generated topology & proxy: reward proxy & generator: graph diffusion & soft objective (cost) \\
\approwspace
Multi-Agent Architecture Search via Agentic Supernet \citep{zhangMultiagentArchitectureSearch2025} & pre-exec / generate & graph & sampled supernet architecture & metric: quality score & generator: architecture distribution & soft objective (cost) \\
\approwspace
ScoreFlow \citep{wangScoreFlowMasteringLLM2025} & pre-exec / generate & graph & workflow generator & preference: score-aware pairs & preference optimization & none \\
\approwspace
FlowReasoner \citep{gaoFlowReasonerReinforcingQueryLevel2025} & pre-exec / generate & graph & operator-library program & metric: reward & RL: meta-controller & soft objective (cost) \\
\approwspace
Workflow-R1 \citep{kongWorkflowR1GroupSubsequence2026} & pre-exec / generate & graph & multi-turn workflow text & metric: grouped rewards & RL: think-act subsequences & none \\
\approwspace
AutoFlow \citep{liAutoFlowAutomatedWorkflow2024} & pre-exec / generate & graph & natural-language DSL & verifier: interpreter feedback & hybrid: generation + execution feedback & none \\
\approwspace
WorkflowLLM \citep{fanWorkflowLLMEnhancingWorkflow2024} & pre-exec / generate & graph & workflow code & supervision: workflow labels & supervised learning: finetuning & none \\
\approwspace
RobustFlow \citep{xuRobustFlowRobustAgentic2025} & pre-exec / generate & graph & robustness-aware generator & preference: clustered preferences & preference optimization & none \\
\approwspace
ComfyUI-R1 \citep{xuComfyUIR1ExploringReasoning2025} & pre-exec / generate & graph & constrained node schema & verifier: validity + reward & hybrid: reasoning training & none \\
\midrule
\multicolumn{7}{@{}l}{\textit{Hybrid drafting and in-execution refinement}}\\
\midrule
AutoAgents \citep{chenAutoAgentsFrameworkAutomatic2024} & pre-exec $\rightarrow$ in-exec / edit & joint & drafted team + plan & trace: runtime feedback & hybrid: draft + observer refinement & none \\
\midrule
\multicolumn{7}{@{}l}{\textit{In-execution editing and adaptation}}\\
\midrule
DyFlow \citep{wang2025dyflow} & in-exec / edit & joint & designer--executor workflow & trace: intermediate feedback & hybrid: online planning + execution & none \\
\approwspace
AgentConductor \citep{Wang2026AgentConductorTE} & in-exec / edit & graph & YAML / DAG topology & verifier: validity + execution & RL: topology revision & hard constraint (budget) \\
\approwspace
Aime \citep{shiAimeFullyAutonomousMultiAgent2025} & in-exec / edit & graph & planner + dynamic actor graph & trace: runtime outcomes & controller: actor instantiation & none \\
\approwspace
AOrchestra \citep{ruanAOrchestraAutomatingSubAgent2026} & in-exec / edit & joint & orchestration state & metric: outcome & controller: sub-agent creation & soft objective (cost) \\
\approwspace
MetaGen \citep{wangMetaGenSelfEvolvingRoles2026} & in-exec / edit & joint & dynamic role pool + graph & trace: running feedback & repair/edit: training-free evolution & soft objective (cost) \\
\approwspace
ProAgent \citep{yeProAgentRoboticProcess2023} & in-exec / edit & graph & structured JSON process graph & verifier: tests & repair/edit: incremental repair & none \\
\approwspace
Flow \citep{niuFlowModularizedAgentic2025} & in-exec / edit & graph & activity-on-vertex graph & trace: intermediate execution & repair/edit: graph refinement & none \\
\approwspace
EvoFlow \citep{zhangEvoFlowEvolvingDiverse2025} & in-exec / edit & graph & evolving workflow population & metric: online fitness & search: online evolution & none \\
\approwspace
DebFlow \citep{suDebFlowAutomatingAgent2025} & in-exec / edit & graph & debated workflow candidates & trace: debate + reflection & repair/edit: debate refinement & none \\
\approwspace
QualityFlow \citep{huQualityFlowAgenticWorkflow2025} & in-exec / edit & graph & fixed scaffold + control actions & verifier: quality checks & controller: submit/clarify/revert & none \\
\bottomrule
\end{longtable}
}

\subsection{Joint optimization of structure and local configuration}
Structure and local configuration are coupled. Changing the topology affects what information each node receives. Changing prompts or tool settings changes what the topology effectively computes. Multi-Agent Design introduces the Multi-Agent System Search (MASS)~\citep{zhouMultiAgentDesignOptimizing2025} framework, which alternates local prompt optimization, topology optimization, and workflow-level prompt optimization under the revised topology. The practical importance of this result is not only that joint optimization helps, but also that a staged schedule can make it stable enough to use in practice.

Learning Multi-Agent Communication from a Graph Modeling Perspective, while not LLM-native, is a useful bridge paper for this survey. It relaxes the discrete communication graph into a continuous parameterization and jointly optimizes graph structure and parameters through gradient-based learning \citep{huLearningMultiAgentCommunication2024}. This differs from discrete search but addresses the same question: communication topology should not be assumed fixed if it materially affects cooperative performance. Maestro~\citep{wang2025maestro} is perhaps the clearest recent statement of the joint view: it alternates graph edits with node-level configuration updates, and it allows both numeric scores and reflective textual feedback to drive the optimization loop. What these papers share is the recognition that prompt tuning alone cannot supply missing structural capabilities such as validation, conditional routing, or intermediate decomposition.

\subsection{Verifiability in static workflow optimization}
Static optimization is particularly compatible with verification because reusable templates can be checked before deployment. This has motivated growing interest in constrained intermediate representations and verifier-driven search. MermaidFlow~\citep{zhengMermaidFlowRedefiningAgentic2025} makes the search space itself more reliable by encoding workflows in a structured, human-readable Mermaid IR, together with static validity checks and safety-constrained evolutionary programming. VFlow~\citep{weiVFlowDiscoveringOptimal2025} integrates stronger external verification directly into the search loop, using syntax, functional correctness, synthesizability, and hardware objectives rather than relying on downstream success alone. In both cases, verification is not added after search; it is part of the optimization process itself.

More broadly, static workflow optimization is most effective when the search space is constrained enough and candidate evaluation is trustworthy enough that offline search can meaningfully separate good structure from bad. When those conditions no longer hold, the central question shifts from how to optimize a reusable template to how much structure should remain plastic at inference time. That shift motivates the dynamic methods reviewed next.

% ============================================================
\section{Dynamic Optimization and Runtime Adaptation}
\label{sec:dynamic}

Dynamic methods determine some part of the workflow at inference time. They are motivated by heterogeneity: different queries require different tools, different reasoning depth, different numbers of agents, and different verification regimes. Their appeal is flexibility, but that flexibility also enlarges the action space, complicates credit assignment, and increases the need for explicit budget guards. Within this family, it is useful to distinguish three increasingly flexible settings: runtime selection over a fixed super-graph, pre-execution workflow generation, and in-execution editing. Table~\ref{tab:dynamic-papers} summarizes representative \textsc{core} dynamic methods with the same comparison card used in Table~\ref{tab:static-papers}.

\subsection{Selection and pruning as the lightest form of runtime adaptation}
The lightest form of dynamism keeps a super-graph fixed and makes run-specific activation or pruning decisions. Adaptive Graph Pruning~\citep{liAdaptiveGraphPruning2025} learns to prune both edges and agents from a complete communication graph using task and agent embeddings, thereby producing task-adaptive sparse topologies. DAGP~\citep{wangDAGPDifficultyAwareGraph2025} further conditions pruning on estimated realized graph difficulty so that easy examples use leaner communication while hard examples retain richer collaboration. AgentDropout~\citep{wangAgentDropoutDynamicAgent2025} similarly removes redundant agents and communication links across rounds to improve the accuracy--cost trade-off. DyLAN~\citep{liuDynamicLLMPoweredAgent2024}, MasRouter~\citep{yueMasRouterLearningRoute2025}, and SkillOrchestra~\citep{wang2026skillorchestra} occupy a similar design point from a routing perspective: they select teams, collaboration modes, models, or tools so that less structure is activated unless the task demands it. Appendix Table~\ref{tab:selection-papers-appendix} summarizes this \textsc{adjacent} family.

This design point has two advantages. First, validity is inherited from the super-graph, which makes optimization easier than unconstrained generation. Second, selective activation often captures a large fraction of the cost savings available from more ambitious dynamic methods. The limitation is expressivity. A pruning policy cannot introduce a missing verifier, a missing tool, or a new decomposition strategy if the super-graph never contained those options. Selection and pruning therefore offer an attractive first layer of dynamism when most of the relevant structure is known in advance and only its activation should vary across inputs.

\subsection{Construct-then-execute: pre-execution workflow generation}
A stronger form of dynamism generates or selects a run-specific workflow before execution begins. Difficulty-Aware Agentic Orchestration for Query-Specific Multi-Agent Workflows~\citep{su2025difficulty} is the lightest variant: it estimates query difficulty and allocates workflow depth, operator choice, and model routing before execution starts, without requiring full graph synthesis from scratch. Assemble Your Crew~\citep{liAssembleYourCrew2025} is more expressive: it autoregressively samples roles and edges to produce a query-conditioned DAG rather than modifying a reusable template. G-Designer~\citep{zhangGDesignerArchitectingMultiagent2025} learns a graph generator with a variational graph autoencoder, while Dynamic Generation of Multi-LLM Agents Communication Topologies with Graph Diffusion Models~\citep{jiang2025dynamic} synthesizes sparse communication topologies through proxy-guided discrete diffusion and multi-objective reward guidance. MaAS~\citep{zhangMultiagentArchitectureSearch2025} learns a query-conditioned distribution over architectures through an agentic supernet. These methods differ in parameterization, but they share the same goal: move from one-size-fits-all collaboration patterns to structures that reflect task complexity and cost sensitivity for a particular run.

Preference and reinforcement learning offer two other routes to pre-execution generation. ScoreFlow~\citep{wangScoreFlowMasteringLLM2025} executes sampled workflows, converts numeric scores into score-aware preferences, and trains a workflow generator with a direct preference optimization (DPO)-like objective that preserves the magnitude information in the scores. FlowReasoner~\citep{gaoFlowReasonerReinforcingQueryLevel2025} trains a query-level meta-agent with reinforcement learning so that a new workflow is generated for each query from an operator library. Workflow-R1~\citep{kongWorkflowR1GroupSubsequence2026} reframes workflow construction as a multi-turn decision process and aligns reinforcement learning with grouped think-act subsequences rather than with a single monolithic generation step. These papers collectively show that once workflow generation is treated as a policy problem, score design and action granularity become central design choices.

Data-centric workflow generation provides a complementary approach. AutoFlow~\citep{liAutoFlowAutomatedWorkflow2024} uses a lightweight workflow DSL and an interpreter model to generate and score workflows from natural-language requirements. WorkflowLLM~\citep{fanWorkflowLLMEnhancingWorkflow2024} instead constructs a large workflow corpus from real automation artifacts and fine-tunes a model to emit workflow code directly. RobustFlow~\citep{xuRobustFlowRobustAgentic2025} adds a robustness objective so that semantically equivalent queries map to more consistent structures. In a more tightly constrained setting, ComfyUI-R1~\citep{xuComfyUIR1ExploringReasoning2025} shows that reasoning-oriented workflow generation can be highly effective when the schema is strict and executability is easy to validate. Although these models are trained offline, they are dynamic in our taxonomy because they produce new realized graphs at inference time.

AutoAgents~\citep{chenAutoAgentsFrameworkAutomatic2024} occupies a hybrid point in this design space. It drafts both team composition and execution plan before execution, then uses observer agents and execution-time refinement to revise them online, blurring the boundary between construct-then-execute and in-execution editing. Taken together, these methods move structural choice to just before execution: it is more expressive than subgraph selection, yet typically easier to validate than fully interleaved runtime editing.

\subsection{In-execution editing: interleaving execution with structural change}
The most flexible dynamic methods treat structural change as a first-class runtime action. DyFlow~\citep{wang2025dyflow} interleaves a designer with an executor, using intermediate feedback to revise sub-goals and dynamically choose operators as execution unfolds. AgentConductor~\citep{Wang2026AgentConductorTE} generates a YAML topology, executes it, then regenerates the topology for the same problem based on validity, code-execution, and cost feedback until success or budget exhaustion. Aime~\citep{shiAimeFullyAutonomousMultiAgent2025} uses a dynamic planner, an actor factory, and a progress manager so that specialized agents can be instantiated and coordinated on demand during long-horizon execution. AOrchestra~\citep{ruanAOrchestraAutomatingSubAgent2026} turns sub-agent creation itself into a callable action, making dynamic specialization part of the orchestration policy rather than a preprocessing step.

Other methods focus on training-free evolution during inference. MetaGen~\citep{wangMetaGenSelfEvolvingRoles2026} initializes a task-adaptive graph and then revises both roles and edges based on lightweight feedback such as contradictions, failures, and cost signals. ProAgent~\citep{yeProAgentRoboticProcess2023}, framed as agentic process automation, constructs workflows in structured JSON and repairs them incrementally through testing-on-constructing. Flow~\citep{niuFlowModularizedAgentic2025} revises an activity-on-vertex graph during execution as new intermediate results arrive. QualityFlow~\citep{huQualityFlowAgenticWorkflow2025} is more conservative: it keeps much of the scaffold fixed but lets a quality checker decide whether to submit, clarify, revert, or continue, thereby making control-flow edits conditional on verifier-like judgments.

A final cluster uses online search or debate. EvoFlow~\citep{zhangEvoFlowEvolvingDiverse2025} maintains a diverse population of heterogeneous workflows and evolves them on the fly. DebFlow~\citep{suDebFlowAutomatingAgent2025} uses multi-agent debate and reflection to improve workflow creation under limited resources. These methods are especially useful when multiple structurally distinct solutions may be reasonable and when diversity helps hedge against search myopia.

Taken together, dynamic methods form a spectrum of structural plasticity, from lightweight subgraph selection to pre-execution generation and full in-execution editing. The relevant design question is not whether more flexibility is intrinsically better, but what minimum plasticity the task actually requires. Moving rightward along this spectrum increases expressivity, but it also raises the burden of validation, stopping, and budget control. Section~\ref{sec:synthesis} returns to this trade-off from a design perspective.

% ============================================================
\section{Feedback Signals and Update Mechanisms}
\label{sec:signals}

The same structural search space can behave very differently depending on the feedback/evidence used to accept, reject, rank, or revise candidate structures. In practice, feedback signal and update mechanism are tightly coupled, so it is useful to discuss them together. A given paper can also use multiple signals at once; throughout this section, categories refer to the primary feedback/evidence used for the main structural decision rather than to an exhaustive list of every signal appearing in the method.

\subsection{Metric- and score-driven optimization}
The most direct signal is a scalar metric such as success, accuracy, F1, pass@k, or an application-specific reward. This signal drives black-box search in AFlow~\citep{zhangAFlowAutomatingAgentic}, ADAS~\citep{huAutomatedDesignAgentic2025}, Evolutionary Generation of Multi-Agent Systems~\citep{huEvolutionaryGenerationMultiAgent2026}, SEW\citep{liuSEWSelfEvolvingAgentic2025}, and VFlow\citep{weiVFlowDiscoveringOptimal2025}. It also underlies query-conditioned generators such as Difficulty-Aware Agentic Orchestration~\citep{su2025difficulty}, FlowReasoner~\citep{gaoFlowReasonerReinforcingQueryLevel2025}, and MaAS~\citep{zhangMultiagentArchitectureSearch2025}, where performance and cost jointly define the reward or routing objective used to train a controller. Scalar metrics are appealing because they align directly with the deployment objective, but they are often sparse, noisy, and expensive. Search efficiency, therefore, depends heavily on the quality of the evaluator and on the granularity of the action space.

\subsection{Verifier-driven optimization}
Verifier signals occupy a special place in workflow optimization because they can be used as hard constraints, intermediate checkpoints, or dense reward components. In program synthesis and software tasks, unit tests and executability checks provide strong supervision for workflow repair \citep{huQualityFlowAgenticWorkflow2025,yeProAgentRoboticProcess2023}. In structured workflow synthesis, format validity and schema compliance can be checked before expensive downstream execution \citep{zhengMermaidFlowRedefiningAgentic2025,xuComfyUIR1ExploringReasoning2025}. AgentConductor~\citep{Wang2026AgentConductorTE} combines validity, execution, and cost feedback during topology revision, and VFlow~\citep{weiVFlowDiscoveringOptimal2025} illustrates the value of multi-level verification particularly well: syntax, functional correctness, synthesizability, and hardware objectives are all measured separately and fed back into search.

Verifiers do more than improve quality. They also shape the feasible edit space. A workflow language with strong static checks enables more aggressive generation or mutation, since invalid candidates can be rejected cheaply. The main limitation is that verifiers are imperfect or expensive themselves. A weak verifier can reward brittle hacks, while an overly costly verifier can dominate the optimization budget.

\subsection{Preference and ranking signals}
Preference signals compare workflows or traces instead of assigning each one an independent scalar reward. ScoreFlow~\citep{wangScoreFlowMasteringLLM2025} constructs score-aware preference pairs from executed workflows and uses them to stabilize gradient-based optimization of a generator. RobustFlow~\citep{xuRobustFlowRobustAgentic2025} builds preferences from semantic clusters and self-consistency, preferring structures that remain stable across paraphrases. Optima~\citep{chenOptimaOptimizingEffectiveness2025} similarly uses ranking and selection over generated multi-agent trajectories before fine-tuning. Preference learning is attractive when absolute rewards are noisy, but relative comparisons are reliable, or when multiple aspects, such as accuracy and cost, must be combined in a ranking rule.

\subsection{Trace-derived textual feedback}
Textual feedback extracted from traces is increasingly used as a higher-bandwidth signal than scalar reward alone. GEPA~\citep{agrawalGEPAReflectivePrompt2025} is the clearest prompt-level example: it reflects on successful and failed trajectories in language and uses those reflections to propose prompt edits. MetaGen~\citep{wangMetaGenSelfEvolvingRoles2026} updates role descriptions and graph edits using contradiction- and failure-oriented feedback from logs. DebFlow~\citep{suDebFlowAutomatingAgent2025} uses debate and reflection to refine workflows. Maestro~\citep{wang2025maestro} explicitly combines numeric scores with reflective text so that graph edits can respond to semantically meaningful failure descriptions rather than only to outcome deltas. SkillOrchestra~\citep{wang2026skillorchestra} occupies a related boundary case: it extracts fine-grained skill abstractions from traces and uses them to guide future routing decisions.

The strength of textual feedback is semantic richness. It can suggest why a workflow failed and what structural change might help. Its weakness is that textual critiques can drift, hallucinate, or overfit to anecdotal traces. In the strongest systems, textual feedback acts as a proposal mechanism, while metrics or verifiers decide whether the proposal should survive.

\subsection{Matching signals to algorithms}
Across the surveyed literature, certain signal--algorithm pairings recur for understandable methodological reasons. Search works best when evaluators are trusted and the action space is discrete but structured, as in typed operator graphs, code templates, or constrained graph IRs. Reinforcement learning is natural when workflow generation is genuinely sequential, as in FlowReasoner~\citep{gaoFlowReasonerReinforcingQueryLevel2025} or Workflow-R1~\citep{kongWorkflowR1GroupSubsequence2026}, but it places great pressure on reward design and exploration. Supervised learning is attractive when workflow artifacts exist at scale, as in WorkflowLLM~\citep{fanWorkflowLLMEnhancingWorkflow2024} or Assemble Your Crew~\citep{liAssembleYourCrew2025}. Training-free adaptation becomes plausible when the edit space is small, the validator is strong, and trace feedback is informative, which explains the recent interest in MetaGen~\citep{wangMetaGenSelfEvolvingRoles2026}, QualityFlow~\citep{huQualityFlowAgenticWorkflow2025}, DyFlow~\citep{wang2025dyflow}, and Aime~\citep{shiAimeFullyAutonomousMultiAgent2025}-style systems.

A useful cross-paper pattern is that the signal often determines the safe action granularity. Strong verifiers support more aggressive graph mutation because invalid candidates can be rejected cheaply. Preference data stabilizes generation when scalar rewards are noisy but rankings are reliable. Textual feedback carries richer causal hints but is safest when coupled to an external validator. RL is most compelling when workflow construction is inherently sequential rather than merely formatted as a sequence. In other words, the literature is not merely choosing algorithms; it is choosing what evidence is trusted enough to justify structural change. This is also why evaluation must report not just final outcomes, but the signals and guards that authorized those changes.

% ============================================================
\section{Evaluation and Reporting for Workflow Optimization}
\label{sec:eval}

Across the surveyed literature, workflow structure is treated as a first-class design object during method development much more often than it is treated as a first-class output during evaluation. For workflow research to mature, the graph itself must become part of what is reported and compared.

\begin{table}[!t]
\centering
\footnotesize
\caption{Representative evaluation assets, grouped by the role they play in workflow optimization research.}
\label{tab:eval-assets}
\begin{tabular}{@{}L{0.20\linewidth}L{0.29\linewidth}L{0.18\linewidth}L{0.11\linewidth}L{0.18\linewidth}@{}}
\toprule
Role in workflow research & Representative resources & Main artifact or evidence & Structure-aware? & Best used for \\
\midrule
Workflow-generation benchmarks & WorFBench~\citep{qiaoBenchmarkingAgenticWorkflow2025}, FlowBench~\citep{xiaoFlowBenchWorkflow2024}, constrained graphical domains such as ComfyUI-R1~\citep{xuComfyUIR1ExploringReasoning2025} & reference workflows, schema-valid graphs, workflow plans & \Yes / \Partial & evaluating structural fidelity, validity, and workflow-level robustness \\
\approwspace
Training corpora and executable validators & WorkflowBench~\citep{fanWorkflowLLMEnhancingWorkflow2024}, HumanEval\citep{chenEvaluatingLargeLanguage2021}, MBPP~\citep{austinProgramSynthesisLarge2021}, GSM8K~\citep{cobbeTrainingVerifiersSolve2021}, MATH~\citep{hendrycksMeasuringMathematicalProblem2021}, DROP~\citep{duaDROPReadingComprehension2019}, HotpotQA~\citep{yangHotpotQADatasetDiverse2018} & workflow code pairs, hidden tests, labeled reasoning tasks & \Partial / \No & supervised workflow generation and controlled validation of local or global design choices \\
\approwspace
Interactive tool-use benchmarks & GAIA~\citep{mialonGAIABenchmarkGeneral2023}, $\tau$-Bench~\citep{yao2025taubench}, $\tau^2$-Bench~\citep{barrestau2BenchEvaluatingConversational2025}, APIBank~\citep{liAPIBankComprehensiveBenchmark2023}, T-Eval~\citep{chenTEvalEvaluatingTool2024}, ToolBench~\citep{xuToolBench2023}, AppWorld~\citep{trivediAppWorldControllableWorld2024}, MCP-Universe~\citep{luoMCPUniverseBenchmarkingLarge2025}, MCP-Bench~\citep{wangMCPBenchBenchmarkingToolUsing2025}, MCP-RADAR~\citep{gaoMCPRADARMultiDimensionalBenchmark2025}, MCPEval~\citep{liuMCPEvalAutomaticMCPbased2025}, LiveMCPBench~\citep{moLiveMCPBenchCanAgents2026}, MCPWorld~\citep{yanMCPWorldUnifiedBenchmarking2025} & long-horizon execution traces under realistic tools & \No & end-to-end orchestration, schema compliance, recovery, and tool-selection quality \\
\approwspace
Software and operations environments & SWE-bench~\citep{jimenezSWEbenchCanLanguage2024}, Terminal-Bench~\citep{merrillTerminalBenchBenchmarkingAgents2026}, SOP-Bench~\citep{nandiSOPBenchComplexIndustrial2026}, AssetOpsBench~\citep{patelAssetOpsBenchBenchmarkingAI2025} & repository edits, terminal traces, procedural execution logs & \No & planning, verification, and recovery under hard constraints \\
\bottomrule
\end{tabular}
\end{table}

\subsection{Treating the workflow as a first-class output}
There are two distinct evaluation goals. The first is \emph{structure-aware evaluation}: did the method generate or select a good workflow or graph? The second is \emph{downstream task validation}: did the resulting system solve the task? Both matter, but they answer different questions. Structure-aware evaluation is most informative when the method explicitly outputs workflows, as in WorFBench~\citep{qiaoBenchmarkingAgenticWorkflow2025} and, to a lesser extent, FlowBench\citep{xiaoFlowBenchWorkflow2024}, or in constrained graphical domains such as ComfyUI-R1~\citep{xuComfyUIR1ExploringReasoning2025}. Downstream validation is indispensable when real execution in tools or environments is the true target, as in GAIA~\citep{mialonGAIABenchmarkGeneral2023}, SWE-bench~\citep{jimenezSWEbenchCanLanguage2024}, Terminal-Bench~\citep{merrillTerminalBenchBenchmarkingAgents2026}, or the broader Model Context Protocol (MCP) benchmark family~\citep{luoMCPUniverseBenchmarkingLarge2025,wangMCPBenchBenchmarkingToolUsing2025}.

A good evaluation protocol should therefore separate the two whenever possible. A workflow generator can output a graph that looks plausible but executes poorly. Conversely, a downstream task can be solved by excessive brute-force compute or by a hidden scaffold that is never analyzed structurally. Reporting only final task success makes it impossible to know which of these actually happened.

\subsection{Effectiveness, efficiency, robustness, and graph-level properties}
Effectiveness metrics remain task-dependent: accuracy, exact match, F1, success rate, pass@k, and domain-specific scores all appear across the literature. Efficiency metrics should be reported alongside them, not after the fact. At a minimum, papers should report prompt and completion tokens, number of LLM calls, number of tool calls, latency or wall-clock time, and monetary cost when priced APIs are used. Cost-per-success is often more informative than raw cost because it normalizes for utility.

Workflow papers should additionally report graph-level properties. For static templates, size, depth, width, critical path length, and communication volume help disentangle ``better structure'' from ``more compute.'' For dynamic methods, edit count, fraction of steps spent editing, and structural variance across repeated sampling or paraphrases are especially informative. RobustFlow's~\citep{xuRobustFlowRobustAgentic2025} node- and graph-level similarity metrics, as well as WorFBench's~\citep{qiaoBenchmarkingAgenticWorkflow2025} subsequence and subgraph matching, show how this style of evaluation can be made concrete.

Robustness should be evaluated under perturbation, not only under nominal execution. The most immediate tests are paraphrase invariance, tool failure injection, information retrieval noise, API or schema drift, and generalization to unseen tools or larger registries. Dynamic methods should also be tested under strict budget caps because the ability to replan can easily degenerate into uncontrolled cost growth. LiveMCPBench~\citep{moLiveMCPBenchCanAgents2026} is a useful step toward drift-sensitive evaluation, but controlled drift protocols remain rare.

\begin{table}[!t]
\centering
\footnotesize
\caption{A proposed minimum reporting protocol for workflow-optimization papers.}\label{tab:reporting}
\begin{tabular}{@{}L{0.25\linewidth}L{0.44\linewidth}L{0.28\linewidth}@{}}
\toprule
Dimension & What should be reported & Why it matters \\
\midrule

Workflow representation & code, DSL, graph IR, schema constraints, executable interpreter, available operators and tools & determines what can be searched, validated, or edited \\
\approwspace
Structural setting & static or dynamic, GDT, GPM, admissible edits, routing policy, stopping rules & clarifies what kind of structural variation the method actually allows \\
\approwspace
Model and tool configuration & base models, decoding settings, tool registry, verifier placement, memory policy & separates workflow effects from backbone or tool effects \\
\approwspace
Offline optimization cost & search budget, number of evaluated candidates, training compute, verifier cost, human annotation if any & distinguishes one-time optimization expense from deployment expense \\
\approwspace
Online inference cost & tokens, LLM calls, tool calls, latency, wall-clock time, dollars, cost-per-success & makes quality--cost trade-offs scientifically comparable \\
\approwspace
Trace statistics & number of rounds, retries, edits, failures, fallbacks, termination causes & exposes hidden control-flow behavior and failure modes \\
\approwspace
Graph-level metrics & node count, depth, width, communication volume, edit count, structural variance & treats the workflow as a first-class output rather than an invisible implementation detail \\
\approwspace
Robustness tests & paraphrases, noisy retrieval, tool failure injection, API drift, unseen tools, strict budget caps & checks whether the workflow policy is stable outside nominal conditions \\
\approwspace
Randomness and evaluation protocol & seeds, repeated runs, sampling temperature, benchmark split, canonicalization rules & reduces variance-induced overclaiming and improves reproducibility \\
\approwspace
Failure analysis & representative traces, verifier disagreements, structural ablations, cases of over-computation & helps identify whether gains come from better structure or just more compute \\
\bottomrule
\end{tabular}
\end{table}

\subsection{A minimum reporting protocol}
Benchmark design itself deserves scrutiny. Rigorous agentic benchmarks need clear task specifications, stable evaluation harnesses, careful budget control, and explicit documentation of what tools and side channels are available \citep{zhuEstablishingBestPractices2025}. Workflow-specific benchmarks additionally need canonicalization rules and structure-comparison procedures; otherwise superficially different but semantically identical workflows will be scored inconsistently. The point is not to force a single canonical graph, but to make evaluation robust enough that structural claims are scientifically meaningful.

The central reporting claim of this survey is therefore simple: workflow papers should not report only \emph{what answer was produced}. They should also report \emph{what workflow was used, how that workflow varied across instances, and what it cost to obtain it}. Without those details, apparent gains are hard to attribute: a method may look stronger because it found a better workflow, because it spent more compute, or because hidden retries and fallback scaffolds were doing the real work.

% ============================================================
\section{Synthesis: Design Trade-offs and Practical Guidance}
\label{sec:synthesis}

The surveyed papers support several broad conclusions, but these should be read as cross-paper empirical patterns rather than as universal laws.

\subsection{When static is enough}
The literature suggests that static optimization is often sufficient when three conditions hold at once: the operator space is constrained enough to search, the evaluator is trustworthy enough to discriminate candidates, and the deployment workload is repetitive enough that a reusable template is worth optimizing. In practice, these conditions often coincide when APIs are stable and verification is strong. In such settings, a well-searched template can dominate ad hoc runtime generation because it is cheaper, easier to debug, and easier to benchmark. This is especially true in domains such as code generation with unit tests or hardware generation with strong external toolchains, where the bottleneck is often not adaptation but disciplined search over a constrained design space~\citep{zhangAFlowAutomatingAgentic,weiVFlowDiscoveringOptimal2025}.

When any of these conditions breaks---for example, under tool drift, environment shift, or strongly heterogeneous task structure---the main weakness of static methods becomes clearer. They can optimize a fixed design very effectively, but they cannot determine that a different design is needed for the current input.

\subsection{When select beats generate, and when edit is unavoidable}
Across the dynamic literature, a useful design rule emerges: different forms of runtime adaptation are justified by different kinds of task heterogeneity. If most instances fall within a known motif library and differ mainly in difficulty, communication budget, or required model strength, then runtime selection or pruning is often the right amount of dynamism. Adaptive Graph Pruning~\citep{liAdaptiveGraphPruning2025}, DAGP~\citep{wangDAGPDifficultyAwareGraph2025}, AgentDropout~\citep{wangAgentDropoutDynamicAgent2025}, and difficulty-aware orchestration~\citep{su2025difficulty} all suggest that substantial savings can come from activating smaller subgraphs or leaner operator allocations on easier instances, without paying the full complexity cost of graph synthesis. In such cases, selection is attractive not only because it is cheaper, but also because it inherits validity from a fixed super-graph and is therefore easier to constrain and debug.

Pre-execution generation becomes preferable when task heterogeneity is larger, and the right structure varies in ways that local prompt tuning or subgraph activation cannot absorb. Methods such as Assemble Your Crew~\citep{liAssembleYourCrew2025}, G-Designer~\citep{zhangGDesignerArchitectingMultiagent2025}, FlowReasoner~\citep{gaoFlowReasonerReinforcingQueryLevel2025}, and Workflow-R1~\citep{kongWorkflowR1GroupSubsequence2026} are strongest precisely when different queries call for genuinely different decompositions, communication patterns, or operator sequences rather than merely larger or smaller versions of the same scaffold. The advantage is greater structural expressivity; the cost is a larger search or generation space, greater dependence on workflow validity, and harder structural credit assignment.

In-execution editing becomes necessary when the environment is interactive enough that partial execution reveals information a one-shot plan could not know in advance. DyFlow~\citep{wang2025dyflow}, AgentConductor~\citep{Wang2026AgentConductorTE}, Aime~\citep{shiAimeFullyAutonomousMultiAgent2025}, AOrchestra~\citep{ruanAOrchestraAutomatingSubAgent2026}, and MetaGen~\citep{wangMetaGenSelfEvolvingRoles2026} are most compelling in settings where plans must survive tool failures, unexpected intermediate outcomes, or evolving task state, and where structural revision is therefore part of solving the task rather than an optional refinement. Moving from selection to generation to editing, therefore, increases flexibility but also puts greater pressure on validity checks, stopping policies, and budget controls.

A practical implication is that dynamic methods should be selected based on the minimum structural plasticity required by the workload. Selection is often the right first step when tasks vary mostly in difficulty or communication budget. Generation becomes worthwhile when different tasks require genuinely different workflow structures. Editing is best reserved for environments where runtime observations materially change what the workflow ought to do next. The cost of moving rightward along this spectrum is not just additional computation, but also additional system-level overhead: explicit guards on tokens, tool calls, and wall-clock time; stopping rules tied to confidence or verifier acceptance; fallback behavior for tool failures; and loop-avoidance mechanisms such as repeated-state detection or edit caps. These are not minor engineering details. They are part of the workflow policy itself and should be evaluated and reported as such~\citep{wang2025dyflow,Wang2026AgentConductorTE,shiAimeFullyAutonomousMultiAgent2025,yeProAgentRoboticProcess2023,huQualityFlowAgenticWorkflow2025}.

\subsection{When graph optimization matters more than prompt tuning}
Prompt optimization is valuable, but in the surveyed papers, it does not replace missing structure. If a failure is caused by missing verification, redundant communication, insufficient decomposition, or incorrect control flow, graph-level optimization is usually the higher-leverage intervention. Multi-Agent Design~\citep{zhouMultiAgentDesignOptimizing2025} and Maestro~\citep{wang2025maestro} make this point directly by showing that prompt and topology updates should inform each other rather than compete for credit. A useful practical heuristic is to inspect traces before optimizing prompts: if the error arises because the wrong node executed, the right node never existed, or the information path was wrong, a better prompt is unlikely to be the real fix.

\subsection{Where verifiers pay off most}
Across the surveyed papers, verifiers deliver the largest returns when they are both cheap and semantically meaningful. Unit tests, schema checks, executability checks, and synthesizability checks are especially valuable because they provide dense, actionable feedback on whether the workflow is heading in the right direction \citep{huQualityFlowAgenticWorkflow2025,yeProAgentRoboticProcess2023,zhengMermaidFlowRedefiningAgentic2025,weiVFlowDiscoveringOptimal2025}. Their value is smaller when they are weak proxies for downstream success or so expensive that they dominate the optimization budget. For example, in a code-generation workflow, running unit tests after a candidate program is produced but before expensive self-debugging or tool-intensive downstream steps can efficiently prune unproductive branches. By contrast, inserting an expensive semantic verifier after every minor retrieval or formatting step may add substantial latency without materially improving final task success. The design question is therefore not merely whether to use a verifier, but where to place it and how often to invoke it.

\subsection{Baselines, hybrids, and a practical recipe}
Stronger structure should be compared against stronger baselines. Framework papers matter here not only as infrastructure, but also because they make reusable structural abstractions explicit. AutoGen~\citep{wuAutoGenEnablingNextGen2023} provides a general conversation-programming scaffold for multi-agent coordination and tool use, while CAMEL~\citep{liCAMELCommunicativeAgents2023} offers an early role-conditioned interaction template for autonomous cooperation. OpenHands~\citep{wangOpenHandsSoftwareAgent2025} anchors software-agent workflows in a stable execution loop with realistic tools. A concrete bridge from scaffold design to workflow optimization is given by MacNet~\citep{Qian2024ScalingLM}, which elevates DAG-based collaboration to a first-class design object; later topology-optimization work such as AgentConductor~\citep{Wang2026AgentConductorTE} can be read as optimizing this design space further by dynamically revising layered topologies under validity, execution, and budget feedback. EvoAgentX~\citep{wang2025evoagentx} provides a complementary example in which the interface between framework and optimizer becomes explicit, integrating workflow generation, execution, and evolutionary refinement within a single open-source platform. At the same time, OneFlow~\citep{xuRethinkingValueMultiAgent2026} offers a useful counterpoint: it argues that some gains attributed to multi-agent workflows can be reproduced by a strong single-agent simulator when roles share the same backbone and can reuse context efficiently. Appendix Table~\ref{tab:framework-papers-appendix} summarizes this \textsc{background} landscape of frameworks and strong baselines.

A practical hybrid recipe follows from the survey. Begin with a constrained static scaffold or a small operator library, and use node-level compilation or prompt optimization to establish a competent baseline. Add graph-level search only when trace analysis reveals structural failure modes rather than purely local instruction errors. Under heterogeneous deployment conditions, prefer runtime selection or pruning before full workflow generation. Reserve in-execution editing for settings with substantial environmental uncertainty, where partial execution reveals information that cannot be anticipated in a one-shot plan. Finally, once a capable design has been found, compress or prune communication through static sparsification or runtime pruning \citep{zhangCutTheCrap2025,liAdaptiveGraphPruning2025,wangAgentDropoutDynamicAgent2025}. This recipe is not universal, but it matches a recurring empirical pattern across the surveyed systems.

% ============================================================
\section{Open Problems and Future Directions}
\label{sec:future}

\paragraph{Structural credit assignment}
The hardest methodological problem remains credit assignment for structural decisions. It is still difficult to determine whether a gain came from a new edge, a new verifier, a changed role prompt, or simply more compute. Better counterfactual replay, ablation-efficient estimators, and critics that operate on traces and graphs together are needed.

\paragraph{Expressivity versus verifiability}
Expressive workflows with loops, dynamic agent creation, and rich conditionals are attractive, but they are difficult to compare and hard to validate statically. Constrained IRs improve executability and reproducibility, but they may exclude the very solutions that make dynamic workflows powerful. Designing representations that balance expressivity and verifiability remains one of the clearest research tensions in this area.

\paragraph{Continual adaptation under tool and environment drift}
APIs, websites, and tool registries change. Static templates need repair. Dynamic generators need drift-aware policies. Yet very few papers report adaptation efficiency: how much extra cost is needed to recover performance once the environment shifts. Live registries are beginning to expose this problem, but controlled drift benchmarks would make it much easier to study rigorously \citep{moLiveMCPBenchCanAgents2026}.

\paragraph{Data and benchmark quality for workflow research}
Workflow optimization depends unusually heavily on the quality of its evaluators and artifacts. If benchmarks leak answers, if reference workflows are inconsistent, or if evaluators reward brittle hacks, optimization will exploit those weaknesses. This is especially acute for structure-aware evaluation, where canonicalization, semantic equivalence, and execution validity all interact. Better curation protocols and benchmark diagnostics are therefore not peripheral concerns; they are central to the validity of the field \citep{zhuEstablishingBestPractices2025}.

\paragraph{Toward theory for workflow optimization}
The field currently borrows tools from search, program synthesis, reinforcement learning, and bi-level optimization, but lacks a clear theory of when dynamic generation is necessary, when static templates are sufficient, and how sample complexity should scale with structural plasticity. Even partial answers would be valuable. They could clarify when runtime edits are worth their additional complexity and when simpler fixed scaffolds should be expected to win.

% ============================================================
\section{Conclusion}
\label{sec:conclusion}
This survey conceptualizes agentic systems as executable workflows and reviews methods for optimizing them. The fundamental taxonomy distinguishes static approaches, which commit to reusable templates prior to deployment, from dynamic methods that instantiate, select, or modify workflow structures conditioned on the current input or runtime feedback. This perspective unifies disparate techniques, such as prompt compilation, topology pruning, workflow generation, and runtime editing within a coherent quality–cost analytical framework. 

We argue that workflow optimization is best understood through three linked objects: the reusable template, the run-specific realized graph, and the execution trace. We further refine the static--dynamic distinction using graph determination time and graph plasticity mode, which helps organize gray cases such as offline-trained workflow generators and runtime subgraph selection. In the literature, metric-based search, verifier-guided optimization, preference learning, and trace-derived textual feedback can all be viewed as ways to learn better local policies, better structures, or both.

Under this view, workflow structure should be elevated from an implementation detail to a first-class design object, recognizing its direct impact on system capability, operational cost, reliability, and scientific comparability. This survey provides the conceptual foundations and analytical apparatus necessary to rigorously analyze, optimize, and evaluate these structures.

% ============================================================
\bibliography{main}
\bibliographystyle{tmlr}

% ============================================================
\appendix
\section{Appendix}
\subsection{Supporting Tables}
\label{app:tables}

This appendix collects supporting material that is useful for completeness but would interrupt the main workflow-centered argument if kept in the body. It first provides catalog tables for node-level prompt optimizers, adjacent routing/pruning methods, and background frameworks. It then records the classification-card convention used throughout the survey.

\begin{table}[htbp]
\centering
\footnotesize
\caption{Node-level prompt optimization methods commonly used inside larger workflows.}
\label{tab:prompt-papers-appendix}
\begin{tabular}{@{}L{0.22\linewidth}L{0.18\linewidth}L{0.18\linewidth}L{0.18\linewidth}L{0.18\linewidth}@{}}
\toprule
Paper & What is optimized & Feedback signal & Method & Why it matters for workflows \\
\midrule
DSPy \citep{khattabDSPyCompilingDeclarative2023} & module prompts and demonstrations & user-defined task metric & compiler / teleprompting & direct local-node optimization in a fixed pipeline \\
\approwspace
Large Language Models as Optimizers \citep{yangLargeLanguageModels2024} & prompts or instructions & external scalar score & LLM-as-optimizer search & black-box optimizer for workflow-local instructions \\
\approwspace
EvoPrompt \citep{guoEvoPromptConnectingLLMs2025} & prompt population & development metrics & evolutionary search & node-level instruction search without graph change \\
\approwspace
CAPO \citep{zehleCAPOCostAwarePrompt2025} & instructions and few-shot content & quality + token-aware cost & cost-aware evolution & makes node tuning sensitive to deployment cost \\
\approwspace
GEPA \citep{agrawalGEPAReflectivePrompt2025} & prompts inside a larger system & task outcomes + reflection & reflective evolution & example of trace-derived text updating workflow-local prompts \\
\bottomrule
\end{tabular}
\end{table}

\begin{table}[tbp]
\centering
\footnotesize
\caption{\textsc{adjacent} methods for selection, pruning, sparsification, and routing.}
\label{tab:selection-papers-appendix}
\begin{tabular}{@{}L{0.22\linewidth}L{0.13\linewidth}L{0.19\linewidth}L{0.18\linewidth}L{0.20\linewidth}@{}}
\toprule
Paper & Setting & What changes at inference time & Signal / method & Why it is adjacent to workflow optimization \\
\midrule
Cut the Crap \citep{zhangCutTheCrap2025} & offline compression & communication edges and message passing & importance proxies + task quality & post-hoc communication sparsification of a discovered scaffold \\
\approwspace
Adaptive Graph Pruning \citep{liAdaptiveGraphPruning2025} & runtime select & agents and communication edges & joint node/edge pruning & learns sparse task-adaptive subgraphs \\
\approwspace
AgentDropout \citep{wangAgentDropoutDynamicAgent2025} & runtime select & redundant agents and links & utility versus token cost & dynamically reduces collaboration overhead \\
\approwspace
DAGP \citep{wangDAGPDifficultyAwareGraph2025} & runtime select & difficulty-conditioned edges & difficulty estimation + task reward & allocates denser communication only when needed \\
\approwspace
DyLAN \citep{liuDynamicLLMPoweredAgent2024} & runtime select & agent team and collaboration pattern & preliminary trials + agent importance & team selection before or during collaboration \\
\approwspace
MasRouter \citep{yueMasRouterLearningRoute2025} & runtime route & collaboration mode, role allocation, model routing & RL under quality--cost objective & routes across multi-agent designs rather than emitting one graph artifact \\
\approwspace
SkillOrchestra \citep{wang2026skillorchestra} & runtime route & agent/model/tool choice & trace-derived skill transfer & makes workflow routing more interpretable and reusable \\
\bottomrule
\end{tabular}
\end{table}

\begin{table}[tbp]
\centering
\footnotesize
\caption{\textsc{background} frameworks, scaffolds, and strong baselines that frequently anchor workflow experiments.}\label{tab:framework-papers-appendix}
\begin{tabular}{@{}L{0.18\linewidth}L{0.24\linewidth}L{0.24\linewidth}L{0.24\linewidth}@{}}
\toprule
Paper & Workflow abstraction & Main contribution & Why it matters in this survey \\
\midrule
CAMEL \citep{liCAMELCommunicativeAgents2023} & role-conditioned multi-agent dialogue & early scalable role-playing template for autonomous cooperation & widely used baseline interaction scaffold \\
\approwspace
AutoGen \citep{wuAutoGenEnablingNextGen2023} & conversation programming over agents and tools & generic infrastructure for multi-agent conversations and tools & reusable scaffold later optimized by search or runtime control \\
\approwspace
EvoAgentX \citep{wang2025evoagentx} & layered framework for workflow generation, execution, and evolution & open-source platform integrating generation and evolutionary refinement & makes the workflow life cycle explicit in one system \\
\approwspace
Scaling Large-Language-Model-based Multi-Agent Collaboration \citep{Qian2024ScalingLM} & DAG-based collaboration network (MacNet) & studies how topology and scale affect performance & motivates topology as a meaningful design variable \\
\approwspace
OpenHands Software Agent SDK \citep{wangOpenHandsSoftwareAgent2025} & software-agent execution loop with terminal and repository tools & open platform for SWE-style agents & important scaffold for workflow experiments in software settings \\
\approwspace
OneFlow \citep{xuRethinkingValueMultiAgent2026} & single-agent simulator over a role-conditioned graph & shows that many homogeneous multi-agent workflows can be simulated by one LLM & strong baseline against overclaiming multi-agent gains \\
\bottomrule
\end{tabular}
\end{table}

\subsection{Literature Classification Convention}
\label{app:card}

For compactness, the main-text tables use a stable classification card rather than paper-specific column designs. Terminology such as ACG template, GDT, GPM, and the scope tags \textsc{core}, \textsc{adjacent}, and \textsc{background} follows Box~\ref{box:glossary}. This section specifies how those terms are operationalized in the comparison tables and makes the coding rules explicit so that the main tables remain internally consistent.

\begin{table}[htbp]
\centering
\footnotesize
\caption{Field definitions and coding rules for the compact classification card used in the main-text tables.}\label{tab:classification-card}
\begin{tabular}{@{}L{0.24\linewidth}L{0.70\linewidth}@{}}
\toprule
Field & Meaning in this survey \\
\midrule
Setting & static or dynamic, together with GDT (offline / pre-execution / in-execution) and GPM (none / select / generate / edit) \\
\approwspace
Optimized level & node / graph / joint \\
\approwspace
Representation & code / text / DSL / explicit graph / typed operator graph / constrained IR / orchestration state \\
\approwspace
Feedback / evidence & dominant workflow-relevant evidence used to accept, rank, or revise structures; encode as a coarse tag plus a short instantiation, e.g., metric, verifier, supervision, preference, trace, or proxy \\
\approwspace
Update mechanism & dominant structural update mechanism; encode as a coarse tag plus a short instantiation, e.g., search, generator, controller, supervised learning, RL, preference optimization, repair/edit, continuous relaxation, or hybrid \\
\approwspace
Cost handling & whether cost enters the method and how: none / eval-only / soft objective / hard constraint, optionally annotated with a resource type such as tokens, tool calls, latency, dollars, or budget \\
\approwspace
Reporting rule & when a paper uses multiple evidence types or update mechanisms, the main-text tables report the dominant workflow-relevant entry in each field; a secondary tag may be added only when it is essential to avoid distortion, with finer details discussed in the surrounding text \\
\bottomrule
\end{tabular}
\end{table}

\end{document}